Abstract: This entry introduces the topic of machine learning and provides an overview of its relevance for applied linguistics and language learning. The discussion will focus on giving an introduction to the methods and applications of machine learning in applied linguistics, and will provide references for further study.

**Introduction**

Machine learning is a subfield of computer science that is concerned with developing methods to teach computers to learn to perform different tasks automatically. It aims to achieve this by "training" computers to recognize patterns of information from a large number of examples related to a given task. After this training phase, the machine is then expected to perform the learnt task autonomously.

Let us consider an application any major email provider employs in daily use: identifying spam emails. The task in this case is to classify any incoming email as spam or non-spam. This is done automatically by initially showing the computer thousands of examples of manually classified spam and non-spam emails from the past. The computer then observes the patterns in both the categories (e.g., something that talks about giving you 1 Million dollars is probably a spam) and builds a "model" of spam identification making use of all the patterns that differentiate between spam and non-spam emails. This model is then used in real-time, on every new email in your inbox. Other examples of such machine learning tasks include: translating text from one language to another, a self-driving car, a product recommendation system on an e-commerce website such as amazon.com and so on. Considering these diverse application scenarios, machine learning has become a popular method of developing computer systems in several domains in the past decade, and applied linguistics is no exception.

This entry explores the relevance of machine learning in the context of technological applications for language learning. Starting with a brief introduction to machine learning, the article will focus on the relevance, methods, and uses of machine learning in three broad application areas in language learning and applied linguistics: assessment, support, and analytics.

**Machine learning -an overview**

Machine Learning (ML) refers to the study of methods to teach computers to learn from large collections of examples. This description may seem similar to statistics, and machine learning indeed has something to do with statistical modeling. However, the focus in machine learning is primarily on making better and accurate predictions on unseen data, based on what the machine has learned. While there are several methods of "learning" from data, most of them fall into one of the following three categories based on the nature of the task to be learned:

a) *Supervised learning*, where the machine is given example inputs along with the expected outputs, and the task is to learn a function that maps the input to its output. The output can be categorical (e.g., spam classification) or numeric (e.g., automated essay scoring). Most of the problems described in this article belong to this category.
b) *Unsupervised learning*, where the computer is not given any information about the expected output, and the task is to predict the latent structure of similarities among the input examples, and group them together based on this similarity. Cluster analysis is one example of an unsupervised learning problem (SEE ALSO: Cluster Analysis).
c) *Reinforcement learning*, where the computer program learns a task in a dynamic environment by slowly working towards a goal and getting feedback from the environment in terms of rewards or punishments. Making a robot learn to hold an object, making a computer play games with humans, and building a self-driving car are some examples of reinforcement learning.

To illustrate how machine learning is implemented in real world applications, let us consider a supervised machine learning problem described earlier: email spam classification. If our goal is to classify email text as spam or not spam, a machine learning approach typically involves the following steps:

Step 1: Training data, which refers to the collection of examples the computer has to learn from. In the case of supervised learning, expected outputs are also given as a part of the training data. As mentioned earlier, training data for spam classification refers to a collection of emails assigned "spam" or "not spam" category by human annotators.

Step 2: Feature representation: This refers to the process of converting the training data into a numeric array of "features" or patterns a machine can learn from. For spam classification, this could mean aspects of text such as: length of the email, presence of all uppercase text, number of spelling/grammatical errors, mention of large amounts of money etc.

Step 3: Learning algorithm, which learns a function to predict outputs from given inputs. This function can then be used as a "model" later on unseen, new data. Logistic regression and linear regression are two examples of familiar statistical methods that can be used to "learn" a function to map from features to outputs, which can later be used for

> prediction as well. There are hundreds of learning algorithms and new algorithms are being developed in research. Some popular supervised learning algorithms that one would typically encounter are: naïve bayes, decision trees, support vector machines, and artificial neural networks based learning approaches (called deep learning).
> Step 4: Evaluation: assessing what the machine learnt, by testing its performance on unseen data, using an appropriate measure. For example, in spam classification, an evaluation measure could be the percentage of emails correctly identified as spam or not spam (classification accuracy) . There are different methods to evaluate a machine learning algorithm, depending on the task and the nature of data it uses.

There are well established algorithms to perform different forms of machine learning. An application of machine learning to a common problem (e.g., spam classification) or problems in a specific domain (such as automated scoring of learner writing in applied linguistics) typically involves using some of the existing algorithms and customizing the feature representations based on the nature of the problem, choosing or creating appropriate training data etc. Machine learning is an active ingredient in several technological applications relevant to applied linguistics, and the rest of this article focusses on that. For a mathematical and theoretical understanding of machine learning, refer to Daume III (2012), and for a more application oriented treatment, refer to Witten, Frank, Hall & Pal (2016). For a conceptual overview without the mathematical rigor, refer to Domingos (2015).

**Machine learning and language learning**
In language learning, machine learning is used to solve a wide range of problems such as: automated scoring of non-native responses (in tests such as GRE and TOEFL), computer assisted spoken language tutoring (e.g., elsaspeak.com), mobile based language learning app development (e.g., duolingo.com), language analysis tools such as spelling-grammar checkers (e.g., grammarly.com) and in education data mining (Baker & Yacef, 2009). An awareness of general ideas behind machine learning will be useful for applied linguists to understand and evaluate such applications relevant to language learning.

These applications can be classified into three broad categories:
a) language and content assessment
b) learning support for reading, writing, speaking and listening
c) learner data analytics
In the first two, which typically involve working with and processing human language, machine learning is combined with Natural Language Processing (NLP) techniques (SEE ALSO: Natural Language Processing and Language Learning). Additionally, machine learning is also an internal component in other technologies such as Automatic Speech Recognition (ASR) and Machine Translation (MT) which are also useful in developing technology tools for language learning and related applications.

**a) Machine learning and Assessment**
Machine learning is primarily used in two aspects of language assessment -  in automated scoring and in test creation.

**i) Automated scoring**:
Automated scoring includes reading, writing, listening and speaking assessment.

Machine Learning and Applied Linguistics

**Writing**: Automated scoring of learner writing generally involves extracting information about different linguistic properties of the text and using them to build a machine learning model. This model is then used to score student responses. Software products such as e-rater® by Educational Testing Service (Burstein, Tetreault & Madnani, 2013) and Intelligent Essay Assessor™ (Foltz, Streeter, Lochbaum & Landauer, 2013) by Pearson Education use automated scoring as a part of their computer based testing systems.

In terms of the four steps mentioned in the previous section, developing an automated assessment method for scoring learners' writing in terms of language proficiency and writing quality can be described as follows:

Step 1: Collect a large corpus of student essays which are scored by human graders.
Step 2: Extract linguistic features of the text that are relevant to assess proficiency using NLP. Some such aspects used in software products such as e-rater® and Intelligent Essay Assessor™ are: spelling and grammar errors (e.g., agreement, preposition errors, capitalization etc), style (repetitions of words or phrases), vocabulary usage, sentence variety, discourse coherence and relevance to the provided writing prompts.
Step 3: Build multiple predictive models with these linguistic features as the patterns the algorithms have to learn, and choose the best performing model after checking for validity and consistency.
Step 4: Use the final model to grade new student essays. Existing automated scoring systems work in conjunction with a human grader (i.e., instead of two human graders, there is one human and one machine grader).

Though this is a well established application for machine learning and natural language processing in applied linguistics, it is also actively studied by the research community as well. Different aspects of automated scoring such as creation of freely available corpora with human rater scores, investigating the role of different linguistic features and algorithms, and developing generalizable scoring methods are current topics of research (e.g., Yannakoudakis, Briscoe & Medlock, 2011; Zesch, Wojatzki,  & Scholten-Akoun, 2015). Machine learning has also been utilized for automated scoring of student writing for languages other than English such as German and Estonian (Hancke, 2013; Vajjala & Lõo, 2014).

**Speaking:** The process for automatically scoring spoken responses for their proficiency follows steps similar to writing assessment, except that the relevant "features" of speech need to be extracted. Some such speech features are measures that quantify fluency and disfluencies in speech and quality of pronunciation. Along with these, measures that evaluate the lexico-grammatical accuracy of speech are also used, based on automatically transcribed text version of the speech sample (Zechner, Higgins, Xi and Williamson, 2009; Bhat, Xue & Yoon, 2014). SpeechRater<sup>SM</sup> (Zechner et.al., 2009) is an example of one such software to perform spoken language assessment. When compared with writing assessment, automatic speaking assessment is not as wide spread in real-world use and is not as accurate either. Thus, current research in this direction primarily focuses on improving the prediction accuracy of machine learning models in performing automatic speech scoring. However, some recent research also discusses the processes and issues involved in applying such models in real-word assessment scenarios

Machine Learning and Applied Linguistics

(Evanini, Hauck & Hakuta, 2017) and ways to efficiently combine human scoring with machine scoring (Yoon & Zechner, 2017).

**Reading and Listening:** Assessment of reading comprehension is typically done by scoring student responses to comprehension questions after reading or listening to content on a given topic. While certain forms of questions such as multiple choice, fill in the blank, yes/no questions are relatively easy and straight forward to evaluate for a machine, evaluating free text short answers automatically is a challenging task and machine learning was found to be useful for short answer scoring.

Machine learning is typically used in short answer scoring for learning a scoring or classification function based on the characteristics of learner responses such as word level overlap, text similarity, semantic relatedness with the teacher's expected answer to a given question etc., These can be considered "features" for this task. When the learnt model sees a new student answer in response to a question, it should be able to compare the student answer with the expected answer and score it accordingly as correct/incorrect or according to the grading scale. While short answer scoring systems did not involve machine learning in their original form, there has been a surge in research related to applying machine learning for short answer assessment with the arrival of data based competitions encouraging researchers to work together to solve the problem using a common data and evaluation framework (e.g., The Hewlett Foundation, 2012). Additionally, machine learning is also used to reduce the efforts of human scorers by clustering similar answers together to reduce the number of responses to be graded, i.e., machine assisted human grading rather than fully automated grading (e.g., Basu, Jacobs & Vanderwende, 2013). Burrows, Gurevych & Stein (2015) discuss the state of the art in automated short answer grading, outlining the trends and methods starting from early work on the topic and describing the state of the art.

**ii) Creation of test items:**
Another important area where machine learning can be useful in language assessment is: automated creation of test items. This primarily involves generating different kinds of questions (e.g.,Wh- question, gap fill, multiple choice etc). While these approaches primarily rely on NLP tools for question generation, machine learning is an integral part in the creation of almost all the modern NLP tools used in this process (e.g., part-of-speech tagging, syntactic parsing, word sense disambiguation etc.). Additionally, machine learning is also useful to estimate the quality (Labutov, Basu & Vanderwande, 2015) and difficulty (Beinborn, Zesch & Gurevych, 2014) of questions.

Since all low stakes and high stakes assessments involve decisions (scores or predictions) that affect human users, one needs to inevitably consider issues such as consistency and fairness of machine generated scores. Recent research (Madnani, Loukina, von Davier, Burstein & Cahill, 2017) on integrating psychometric recommendations on quantifying potential biases into automated assessment is a step in that direction.

**b) Machine Learning and Learning Support**
Second area of applied linguistics where machine learning is useful is the development of learning support tools and applications. Learning support tools include various forms of tutoring systems and other tools that provide reading, writing, speaking and listening support and other software such as language games and apps.

Machine Learning and Applied Linguistics

     Writing support tools such as spelling and grammar checkers primarily rely on NLP for identifying and correcting error patterns in learner texts. However, machine learning is increasingly combined with NLP in the development of grammar check/correction both in research (Rozovskaya & Roth, 2014) and in real-life software products in the technology industry (e.g., grammarly.com, deepgrammar.com). Machine learning is primarily used in these tools for two tasks: detection and classification of an errors (e.g., a preposition error versus a determiner error) based on linguistic features, as well as to correct the errors. Along with Machine Translation, machine learning is one of the primary methods used in context sensitive spelling and grammar correction, as it can seen from the reports on grammatical error correction shared tasks organized in NLP community in the recent years (e.g., Ng, Wu, Briscoe, Hadiwinoto, Susanto, & Bryant, 2014). Going beyond spelling and grammar correction, machine learning is also used to build customized writing tutors such as the Research Writing Tutor (Cotos, 2014) which provides automatic feedback to students writing research articles. Machine learning is used in this software to analyse the sentences written by the authors and automatically classify their rhetorical function in the text based on language use.

     Machine learning is a major component in reading support tools such as REAP (Heilman, Zhao, Pino & Eskenazi, 2008) and TextEvaluator$^{SM}$ (Sheehan, K. M., Kostin, I., Napolitano, D., & Flor, M. (2014)which evaluate the linguistic characteristics of a text and give predictions about its reading level and language complexity. Such tools are useful in instruction and support to ensure students read texts they can comprehend. They are becoming increasingly relevant in the wake of the Common Core Standards initiative in the US (Common Core Standards, 2010). As a component in automated speech recognition systems, machine learning is also actively used in pronunciation and listening support systems (Strik, Truong, de Wet & Cucchiarini, 2009).

**c) Machine Learning and Learner Analytics**
Machine learning is also used in other aspects of language learning that are not directly concerned with an analysis of language such as educational data mining and learning analytics. This refers to the collection and analysis of learners' interactions with a computer based tutoring system such as the learner's engagement with the system, exercises done, time taken to complete, time spent reading and re-reading etc.This is then used to create a learning profile of a student, which can be used to predict his/her future performance in the course. The goal of such predictive analytics is to eventually be able to provide individualized learning considering aspects such as learning difficulty/ease and provide specific feedback on next steps in instruction. Additionally, this is also used for providing feedback about various aspects of the course to instructors too. Although initial research in this direction relied on hand crafted rules from domain experts on specific questions, machine learning is now being used in the development of personalized learning solutions for different learning scenarios, including but not limited to massive open online courses (Lan, 2016). It is also used to perform analytics on data such as automatically summarizing course feedback given by students(Luo, Liu & Litman, 2016), to gain insights about teaching and learning.

     In terms of real-world application, mobile based language learning apps such as Duo Lingo (duolingo.com) and Elsa Speak (elsaspeak.com) use machine learning at various stages such as the creation and evaluation of exercises, automatic speech recognition (in speech exercises), adapting to the student, sequencing of lessons, modeling and tracking the learning process, and providing effective, individualized feedback. In such application scenarios where there is a continuous interaction and tracking of the user, the machine learning model, rather than

Machine Learning and Applied Linguistics

undergoing a one time training, is seen as a more dynamic, active learning system, which continuously adapts itself to individual students as they start using the application more and more.

**Conclusion:**
In conclusion, machine learning is used in a wide range of applications in the context of language learning and in different contexts such as reading, writing, speaking, assessment, language tutoring apps, and personalized learning. An understanding of the role of machine learning is important for applied linguists to work with technological tools in language learning. Similarly, an awareness of learning processes, validity of the predictive models and other issues studied in applied linguistics is essential to develop better machine learning systems for these applications. More collaborative work between the computational and the non-computational ends of language learning studies will result in better integration of advances from both ends into language learning applications of the future.

**SEE ALSO:**
Automatic Speech Recognition; Cluster Analysis; Emerging Technologies for Language Learning; Information Retrieval for Reading Tutors; Intelligent Computer-Assisted Language Learning; Learner Corpora; Learner Modeling in Intelligent Computer-Assisted Language Learning; Natural Language Processing and Language Learning; Technology and Language Testing; Text-to-Speech Synthesis in Computer-Assisted Language Learning

Machine Learning and Applied Linguistics

**Suggested Readings:**
1. Bishop, C. M. (2006), Pattern Recognition and Machine Learning, Springer, ISBN 0-387-31073-8
2. Handbook of automated essay evaluation: Current applications and new directions
3. Harrington, P. (2012). Machine learning in action (Vol. 5). Greenwich, CT: Manning.
4. Jurafsky, D., & Martin, J. H. (2000). Speech and Language Processing: An Introduction to Natural Language Processing, Computational Linguistics, and Speech Recognition.